\documentclass[lettersize,journal]{IEEEtran}
\usepackage{amsmath,amsfonts}
\usepackage{algorithmic}
\usepackage{algorithm}
\usepackage{array}
\usepackage[caption=false,font=normalsize,labelfont=sf,textfont=sf]{subfig}
\usepackage{textcomp}
\usepackage{stfloats}
\usepackage{url}
\usepackage{verbatim}
\usepackage{graphicx}
\usepackage{cite}
\usepackage{orcidlink}
\usepackage{booktabs,multirow}
\usepackage{dsfont}

\begin{document}

\title{Hybrid X-Linker: Automated Data Generation and Extreme Multi-label Ranking for Biomedical Entity Linking}

\author{Pedro Ruas\orcidlink{0000-0002-1293-4199}, Fernando Gallego\orcidlink{0000-0001-8053-5707}, Francisco J. Veredas\orcidlink{0000-0003-0739-2505}, Francisco M. Couto\orcidlink{0000-0003-0627-1496}

\thanks{
P. Ruas and F. Couto were with \textit{LASIGE, Faculdade de Ciências, Universidade de Lisboa, 1749-016, Lisbon, Portugal, e-email: psruas@fc.ul.pt}.

F. Gallego and F. Navarro was with \textit{Departamento de Lenguajes y Ciencias de la Computación, Universidad de Málaga, Málaga, Spain} and \textit{with Research Institute of Multilingual Language Technologies, Universidad de Málaga, Málaga, Spain}.
}

\thanks{This work has been submitted to the IEEE for possible publication. Copyright may be transferred without notice, after which this version may no longer be accessible.}

}

\markboth{Preprint submitted to IEEE Transactions on Knowledge and Data Engineering}%
{Ruas \MakeLowercase{\textit{et al.}}: Hybrid X-Linker}


\maketitle

\begin{abstract}
State-of-the-art deep learning entity linking methods rely on extensive human-labelled data, which is costly to acquire. Current datasets are limited in size, leading to inadequate coverage of biomedical concepts and diminished performance when applied to new data. In this work, we propose to automatically generate data to create large-scale training datasets, which allows the exploration of approaches originally developed for the task of extreme multi-label ranking in the biomedical entity linking task. We propose the hybrid X-Linker pipeline that includes different modules to link disease and chemical entity mentions to concepts in the MEDIC and the CTD-Chemical vocabularies, respectively. X-Linker was evaluated on several biomedical datasets: BC5CDR-Disease, BioRED-Disease, NCBI-Disease, BC5CDR-Chemical, BioRED-Chemical, and NLM-Chem, achieving top-1 accuracies of 0.8307, 0.7969, 0.8271, 0.9511, 0.9248, and 0.7895, respectively. X-Linker demonstrated superior performance in three datasets: BC5CDR-Disease, NCBI-Disease, and BioRED-Chemical. In contrast, SapBERT outperformed X-Linker in the remaining three datasets. Both models rely only on the mention string for their operations. The source code of X-Linker and its associated data are publicly available for performing biomedical entity linking without requiring pre-labelled entities with identifiers from specific knowledge organization systems.
\end{abstract}


\section{Introduction}
\IEEEPARstart{E}{ntity Linking} (EL) is the task of linking an entity mention in a given piece of text to an entry in a target Knowledge organization system (KOS), such as an ontology, a knowledge base or graph, a terminology, etc. The entry must accurately represent the meaning of the linked entity. EL is essential in text mining and natural language processing pipelines since it connects text expressed in natural language to semantic, computer-friendly representations. In the biomedical field, substantial amounts of information are captured in clinical notes written in natural language. These notes include entities that require standardization using ontologies like SNOMED-CT or UMLS.

Challenges in the biomedical EL task include name variations (synonyms, acronyms), ambiguity (where the same name can denote different entities)\cite{Shen2015}, and the highly specialised language, which hinders the use of complex resources typically available for general EL approaches, such as Wikipedia. The challenge of ambiguity is illustrated by the entity mention ``iris", which can have several possible meanings: an eye-related anatomical structure, an insect or a plant taxonomic genus, a disease's acronym (immune reconstitution inflammatory syndrome) or a gene. Searching for ``iris" in NCBI-Gene returns multiple homonymous results: ``[Drosophila melanogaster (fruit fly)]" - \texttt{Gene ID: 33290}), ``Iris iris [Tribolium castaneum (red flour beetle)]" - \texttt{Gene ID: 103314968}, and ``Iris iris [Dalotia coriaria]" - \texttt{Gene ID: 135789998}\footnote{Search performed in June 5th, 2024}. Besides, the insufficient coverage of the target KOS results in outdated information and unlinkable entity mentions \cite{Ruas2022}.

Another challenge is that current state-of-the-art approaches resort to supervised, deep-learning-based approaches that require abundant quality annotated data \cite{Shen2015}. Large human-labelled datasets are expensive and hard to build since their creation requires biomedical expertise \cite{Su2019,Sevgili2022}. The performance of the deep learning models is bounded by the information accessed during their training. The applicability will be similarly restricted if most of the datasets are small-scale. To unlock the vast amount of biomedical text available and improve the performance of the task, it is necessary to go beyond supervised approaches trained on limited human-annotated data.

To evaluate the true generalization capability of EL approaches, in \cite{Tutubalina2020} the authors underscored the significance of assessing these methods, in particular supervised ones, on refined test sets. These test sets exclude annotations that are concurrently present in both the training and development sets. The performance of a supervised approach is heavily reliant on the dataset, which does not align with a real setting where these approaches are employed for inference without artificial partitions of the available data into training and test sets.

To mitigate the requirement for costly human-labelled training data, emphasis should be placed on domain-independent approaches trained on domains with large amounts of labelled data and then applied in domains with limited labelled data available \cite{Sevgili2022}. To achieve this objective, distant supervision \cite{Pattisapu2020,Fan2015,Le2019} and zero-shot methods \cite{Liu2021,Zhang2022} have been investigated in the context of the EL task.

Distant supervision consists of generating training data using only a limited amount of human-labelled data, for example, information such as concept names, relations, etc present in a curated KOS \cite{Mintz2009,Le2019}. The entity names and synonyms described in the target KOS can be matched with unlabelled text to label annotations instances \cite{Le2019}. Zero-shot methods focus on generalising an approach to new domains and entities, which were not accessed during the training stage. The key idea is to develop an approach able to link the entities by having only access to the descriptions of the entities belonging to the target domain \cite{Sevgili2022}.

Biomedical KOS typically represent a large number of concepts, however not every concept is represented in the datasets used to evaluate the EL task. The BC5CDR dataset \cite{bc5cdr} includes 4,424 disease annotations, associated with 674 MEDIC vocabulary concepts, and 5,385 chemical annotations associated with 676 CTD-Chemical vocabulary concepts (check Table \ref{tab:eval_datasets}) \cite{Davis2023}, which corresponds to a KOS concept coverage of 5.1 \% and 0.38\% in the dataset, respectively\footnote{MEDIC and CTD-Chemical vocabularies version:Feb 28 2024 10:59 EST}. To develop EL approaches capable of handling a large number of concepts in the target KOS, relying solely on evaluation datasets is insufficient.

In our work, we expand on the idea of distant supervision and zero-shot to build a large-scale training dataset and an EL approach that can link disease and chemical entities without the need for retraining with human-labelled data. To effectively train a deep-learning-based model in the generated large-scale dataset, we frame the EL task as an extreme multi-label ranking (XMR) problem \cite{Chang2020}, where there are a large number of source texts to label as well a large set of target labels and adapt them to the EL task. We explore the hypothesis of applying XMR approaches to the biomedical domain by training models designated by PECOS-EL, jointly with several types of EL approaches that have proven effective in the task. The contributions of this work include:

\begin{itemize}
    \item Development of the PECOS-EL Model and X-Linker pipeline, comprising modular components designed for biomedical KOS to link disease and chemical entities.
    \item Creation of large-scale training datasets featuring automated entity annotations of chemical and disease entities.
    \item Source code publicly available to allow experiment reproducibility and further improvements: \url{https://github.com/lasigeBioTM/X-Linker}
\end{itemize}

\section{Related Work}
In the past decade, varied types of approaches have been proposed to address the problem of EL in the biomedical domain, ranging from heuristics \cite{Dsouza2015} to the most recent deep-learning-based architectures \cite{Liu2021,Yuan2022,Zhang2022}.

Rule-based approaches offer the advantage of bypassing the requirement for a large volume of labelled data, albeit at the cost of performance. For example, \cite{Dsouza2015} proposed a multi-pass sieve approach for EL in clinical records and scientific articles. This work shares some similarities to our work in the sense that it outlines a rule-based pipeline that includes multiple entity processing steps, including string matching the input mentions to the target KOS, abbreviation expansion, identification of composite mentions and several syntactic transformations, such as stemming, hyphenation or dehyphenation, suffixation and the replacement of numbers by their extended form. 

Machine learning-based supervised approaches improve the performance in the task, but require human-labelled data. For instance, TaggerOne \cite{Leaman2016} is a machine learning-based approach that models jointly NER and EL. The EL component is a supervised semantic indexer that generates vectorized representations for both input token and candidate KOS entities and then it assesses the correlation between tokens and target KOS entities using a semi-Markov model.

Supervised approaches work better for domains with plenty of labelled data available, such as the general domain that has Wikipedia. Therefore, more recent approaches to the biomedical EL task focus on deep-learning architectures that require less annotated data.

For instance, BioSyn \cite{Sung2020} focus on learning sparse and dense representations for entities using the synonym marginalization technique. The approach applies an iterative candidate retrieval to maximise the marginal likelihood of the synonyms being present in the top candidates.

The recently proposed BELHD \cite{Garda2024} expands on BioSyn by focusing on homonym entities. The approach replaces homonym entities with a disambiguated version included in the target KOS and then introduces candidate sharing and a new objective function to train the BioSyn model.

BERN2 \cite{Sung2022} shares similarities with our work as it adopts a hybrid approach: initially employing a rule-based module to link entities, then applying the deep-learning BioSYN model for more challenging cases. However, BERN2 is a supervised approach that uses annotations from the training sets of evaluation datasets to fine-tune BioSYN.

Several zero-shot approaches have been proposed to tackle the EL task but these mostly focus on the general domain \cite{Logeswaran2019, Yao2020,Wu2020,Tang2021}. In the biomedical domain, two zero-shot approaches have achieved state-of-the-art performance: SapBERT \cite{Liu2021} and KRISSBERT \cite{Zhang2022}. SapBERT \cite{Liu2021} represents an unsupervised approach with a focus on learning representations for entities within the target KOS. The method involves pretraining a Transformer-based model on UMLS data using a self-alignment objective. Initially, it clusters synonyms of UMLS entries, after which a BERT-based model learns a mapping function between names and their corresponding Concept Unique Identifiers (CUIs). KRISSBERT \cite{Zhang2022} introduces a self-supervised method for EL aimed at mitigating the scarcity of annotated data for model training. The approach generates entity annotations by matching UMLS entity names with unlabelled PubMed documents. Subsequently, it employs contrastive learning to train a contextual encoder. This involves creating positive pairs, where two entity mentions are associated with the same UMLS CUI, and negative pairs, where two entity mentions are linked to different CUIs. The encoder is trained to map mentions of the same entity closer together and mentions of different entities further apart, thereby generating distinct representations for UMLS entities.

Recently, in \cite{Jiang2024} the authors introduced an XMR-based approach for general EL, demonstrating the adaptability of techniques initially developed for XMR to the EL task. There are several key differences between their approach and ours: their entity retriever utilizes beam search, while ours employs a BERT-based matcher (XR-Transformer); their method is applied to datasets annotated with Wikipedia entities, whereas we focus on the biomedical domain. Additionally, we implement a hybrid pipeline that incorporates modules for various types of EL, whereas they solely apply the XMR-based model. We further provide an extensive description of our proposed approach to the EL task.

\section{Methods}

\subsection{Named Entity Linking definition}
Let \( T \) be a text document containing a set of entity mentions \( M = \{m_1, m_2, \ldots, m_n\} \), and \( KOS \) a knowledge organization system including a set of entities or concepts \( E = \{e_1, e_2, \ldots, e_k\} \). The goal of EL is to map each mention \( m_i \in M \) recognized in a given document to its corresponding entity \( e_j \in E \). Ideally, input entity mentions are linked to entities that accurately represent their semantic meaning. 

There are two essential phases in the EL task: 

\begin{itemize}
    \item \textbf{Candidate Generation}: the goal is, for each mention \( m_i \), to generate a set of candidate entities \( C_i \subseteq E \).

    \item \textbf{Candidate ranking and disambiguation}: the goal is to rank the candidate entities \( C_i \) based on their similarity scores \( \text{sim}(m_i, e_j) \). Depending on the approach, the similarity can be calculated based on the features of the individual mentions (\textbf{local approach}), the features of other mentions within the same document (\textbf{global approach}), or a combination of both. The entity \( e^*_i \) with the highest similarity score in the candidates set is selected:
    \[
    e^*_i = \arg\max_{e_j \in C_i} \text{sim}(m_i, e_j)
    \]
\end{itemize}

The final mapping \( \mathcal{M} \) from mentions \( M \) to entities \( E \) is given by:
\[
\mathcal{M}(m_i) = e^*_i \quad \forall m_i \in M
\]

There are various methods for candidate generation and ranking in the EL task. We demonstrate in this work that no single approach is universally optimal for all entities. Instead, a combination of different approaches yields better performance.

\subsection{Entity Linking as a string similarity problem}
The candidate generation is achieved using a string similarity function. One commonly employed string similarity function is the edit distance also designated by Levenshtein Distance. The Levenshtein distance between two given strings represents the minimum number of single-character edits (insertions, deletions, or substitutions) necessary to convert one string into the other. Defining the distance as \( d \), the distance between a given mention \( m_i \in M \) and an entity \( e_i \in E \), the goal is to compute the distance between a mention and every entity and then choose the entity the smallest distance to link the mention to. The limitation is that it relies solely on individual features of the input entity mention, and these features are strictly string-based, lacking consideration of contextual information.

\subsection{Entity Linking as an eXtreme Multilabel Ranking problem: PECOS-EL}
In this work, we investigate framing of the biomedical EL task as an XMR problem, which to the best of our knowledge, has been only recently attempted in the general domain by \cite{Jiang2024}. Given an input entity mention, the goal is to return the most relevant labels or identifiers from a large set of labels included in the target KOS. We used PECOS \cite{yu2022}, a framework originally designed for Information Retrieval approaches. In the EL task, the input mention serves as the text and the set of entities $E$ in the KOS represents the target labels. The PECOS framework encompasses three stages: 

\begin{enumerate}
    \item \textbf{Semantic Label Indexing}: the set of KOS entities $E$ is partitioned into $K$ clusters.

    \item \textbf{Matching}: an entity mention is mapped into relevant clusters through a learned scoring function.
        
    \item \textbf{Ranking}: a ranker assigns scores to the candidate entities present in the matched clusters.
\end{enumerate}

In semantic label indexing, labels/entities from a target KOS are grouped into clusters reducing the search space. Representations for each entity ${z_{e}: e \in E}$ are obtained by aggregating the feature vectors of the training instances associated with the entity. The clustering algorithm maps each entity to a cluster: $c_e \in Cl^E$, where $c_e$ denotes the index of the cluster containing the entity $e$. The clustering is represented by the clustering matrix \( Cl^E \in \{0, 1\}^{E\times K} \) with $E$ representing the entities in the target KOS and $K$ representing the number of entity clusters.

During the matching stage, a general matcher function \( g(x, k) \) determines the relevance between an instance \( x \) (an entity mention) and the $k$-th entity cluster. The top-\( b \) clusters in $Cl$ are identified through \( g_b(x) \):

\[ g_b(x) = \arg \max_{S \subset Cl : |S| = b}  \sum_{k \in S} g(x, k) \]

The function $g_b(x)$ attempts to find the subset $S$ of size $b$ included in $Cl$ that maximises the function $g$ evaluated at each $S$ for a given $x$. The deep text vectorizer is a pre-trained Transformer model, specifically BioBERT. We briefly explored other BERT-based models (BERT, SciBERT, BioBERT, PubMedBERT), but we found the differences to be minimal.

After the matching stage, the ranker \( h(x, e) \) models the relevance between \( x \) and each candidate entity belonging to the clusters previously identified by the matcher function \( g_b(x) \).

We trained two PECOS models for two entity types, 'Disease' and 'Chemical. We further describe the generated training data.

\subsection{Generation of training data with automatic labelling}\label{train_data}
Deep-learning-based approaches that focus on specific tasks usually require a vast amount of human-labelled data, which is scarce in the biomedical domain. The annotation process is a bottleneck in the development of such approaches since it is slow, costly and it requires biomedical expertise. To overcome this obstacle, we generated training datasets for the model PECOS-EL that include automatic annotations obtained from Pubtator3 \cite{Wei2024} and the targets KOS.

\subsubsection{Pubtator3 data}
Pubtator3 is a resource for exploring the biomedical literature present in PubMed\footnote{\url{https://pubmed.ncbi.nlm.nih.gov/}}, providing information retrieval and extraction utilities. Pubtator3 also includes deep-learning-bases tools for entity recognition and linking focusing on six common biomedical entity types: \textit{Gene}, \textit{Chemical}, \textit{Disease}, \textit{CellLine}, \textit{Species}, \textit{Variant}. These entity types are linked, respectively, to the resources: NCBI Gene database, the Medical Subject Headings (MeSH) thesaurus, MeSH, Cellosaurus, NCBI taxonomy, and the NCBI Single Nucleotide Polymorphism (dbSNP) database. The PubTator3 FTP site\footnote{\url{https://ftp.ncbi.nlm.nih.gov/pub/lu/PubTator3/}} provides bulk downloads of all PubMed abstracts and the full texts associated with articles from the PMC Open Access Subset (PMC-OA), as well the respective entity and relation annotations. Pubtator3 applies TaggerOne \cite{Leaman2016} and NLM-Chem \cite{Islamaj2021} to link disease and chemical entities, respectively. The micro-averaged F1-score of Pubtator3 determined in the BioRED dataset is 0.7917 and 0.8192 for disease and chemical entities (data extracted from ``Supplementary Table 3" in \cite{Wei2024}). The latest version of Pubtator3 includes 135,861,884 chemical annotations (file ``chemical2pubtator3.gz") and 154,124,935 disease annotations (file ``disease2pubtator3.gz"). We applied a pre-processing pipeline to convert each Pubtator file into useful training data:

\begin{enumerate}
    \item Document removal: deletion of annotations associated with documents present in the evaluation datasets (see Table \ref{tab:eval_datasets}.
    \item Lowercase the text of each annotation.
    \item Deduplication of annotations.
    \item Removal of obsolete target KOS identifiers and conversion of the identifier into numerical indexes.
    \item Sorting the annotations associated with each KOS identifier according to their frequency in the Pubtator3.0 set of annotations.
\end{enumerate}

\subsubsection{KOS data}
The target KOS consisted of the following curated data retrieved from the Comparative Toxicogenomics Database (CTD) \cite{Davis2023} (MDI Biological Laboratory, Salisbury Cove, Maine, and NC State University, Raleigh, North Carolina, URL: \url{http://ctdbase.org/}): \textit{Disease Vocabulary} (also called \textit{MEDIC})\footnote{Version:Feb 28 2024 10:59 EST}, \textit{Chemical Vocabulary}\footnote{Version:Feb 28 2024 10:59 EST}. Both \textit{MEDIC} and \textit{CTD-Chemical} include entities associated with the respective MeSH identifiers, which allows us to integrate both KOS and Pubtator3 information in a common data space for training.

\subsubsection{Training datasets}
 We generated files with training data for each entity type: ``Chemical" and ``Disease". Each entity type has several dataset versions. Table \ref{tab:train} summarizes the information for all generated files. As a baseline, the training files ``Disease-KOS" and ``Chemical-KOS" only include names and synonyms extracted from the target KOS, more concretely, the \textit{MEDIC} and the \textit{CTD-Chemical} vocabularies. For the \textit{Chemical} training file ``Chemical-All", all the annotations present in the set provided by Pubtator3 of the type 'Chemical' that had a valid MeSH identifier (i.e. an identifier present in \textit{CTD-Chemical} version used in this work) were included. For the ``Disease" training files, given the higher number of available annotations, we generated several versions of the dataset, setting as threshold the number of maximum instances allowed per KOS entity: ``Disease-100", ``Disease-200", ``Disease-300", ``Disease-400". We also generated the file ``Disease-All" including all the Pubtator3 annotations and KB names and synonyms. The structure of the training file includes two columns: a numerical index associated with the respective KOS identifier and the string associated with the annotation or the KOS canonical name or synonym.

\begin{table}[ht]
    \caption{Versions of the generated training files per entity type and respective number of instances. For the 'Disease' training data, the training data included at most 100, 200, 300 and 400 per KOS entity, as well one version with all instances}
    \label{tab:train}
    \centering
    \begin{tabular}{l l l l}
    \toprule
         \textbf{Type} & \textbf{Name} & \textbf{Description} & \textbf{Instances} \\
         \hline
        \multirow{2}{*}{Chemical} & Chemical-KOS & KOS labels & 452,318 \\
        & Chemical-All  & KOS labels + Pubtator (all) & 1,123,842 \\
        \hline
        \multirow{5}{*}{Disease} & Disease-KOS & KOS labels & 89,465 \\
        & Disease-100 & KOS labels + Pubtator (100) & 828,163 \\
        & Disease-200 & KOS labels + Pubtator (200) & 1,402,332 \\
        & Disease-300 & KOS labels + Pubtator (300) & 1,873,901 \\
        & Disease-400 & KOS labels + Pubtator (400) & 2,275,258 \\
        & Disease-All & KOS labels + Pubtator (All) & 9,497,985 \\
        \bottomrule
    \end{tabular}
\end{table}

The training data incorporates alternate names for each entity. Nevertheless, in certain instances, integrating information about the context of the entity into the linking decision can enhance performance.

\subsection{Entity linking as collective coherence maximization problem: Personalized PageRank}
One of the main obstacles in the EL task is the presence of homonym entities, i.e., entities sharing the same string but with highly different meanings \cite{Garda2024}. One way to diminish the impact of such cases is by applying a global approach, which takes into account the document context to perform the linking process. In this type of approach, a given entity mention is linked according to how the other entity mentions present in the same document are linked. We previously demonstrated how the Personalized PageRank (PPR) algorithm can be integrated into such global approach \cite{Lamurias2019,Ruas2020}.

In a given document \( T \), for each entity mention \(m_i \in M \), the approach generates a set of candidate entities \( C_i \subseteq E \). Using these candidates entities, the approach builds a graph disambiguation \( G \), represented as \( G(N, V) \), with \( N \) as the set of nodes in the graph and \( V \) as the set of vertices or edges connecting the nodes. Each node \( n \in N \) corresponds to a pair consisting of an entity mention and its respective KOS candidate. The graph can be described as \( G = \{(m, c) \mid m \in M, c \in C \} \). The edges between candidate nodes are based on the direct edges defined in the target KOS, for instance, on \textit{is-a} relationships. The original PageRank algorithm \cite{Page1999} simulates random walks on a graph, and in each walk, there is a teleport probability \( e \) of going to a random node and a \( 1 - e \) probability of going to a node connected with the current one. In the PPR \cite{Pershina2015} variation, the teleports are always performed to some predefined source node. The stationary distribution resulting from these walks assigns scores or weights to each node in the graph. The PPR algorithm calculates the coherence of each node in the graph \( G \), i.e., how well the node fits into the set of all nodes. The algorithm starts by measuring the pairwise coherence of a source node \( s \) and a target node \( t \):

\[ coherence_{s}(t) = PPR(s \to t) \]

Following the previous approach developed by our group \cite{Lamurias2019}, we enhance the coherence by multiplying it by the information content (IC) of the node \( t \). This adjustment encourages the algorithm to select more specific entries within the KOS at the expense of more general ones:

\[
\text{coherence}_{s}(t) = \text{PPR}(s \to t) \cdot \text{IC}(t)
\]

We opted for the intrinsic definition for IC, in which the IC of a KOS entity \( e \) is given by its frequency in the respective KOS \cite{Couto2019}:

\[
\text{IC}(e) = -\log(p(e))
\]

where \( p \) is the probability of the entity \( e \) and is represented as 

\[
p(e) = \frac{\text{Desc}(e) + 1}{|E|}
\]

Where $Desc$ correspond to the number of child entities or direct descendants of the entity $e$ in the structure of the target KOS, and $|E|$ is the set of every entity represented in the target KOS.

After calculating all the pairwise coherences for node \( s \), the global coherence of \( t \) is given by the sum of its coherence with each source node \( s \):

\[
coherence (t) = \sum_{s \in G}  coherence_{s}(t)
\]

One drawback of this approach is its vulnerability to noise propagation. In certain scenarios, there might be multiple entity mentions with ``imperfect" candidate lists, meaning the list either lacks the correct candidate or contains candidates that are highly unrelated to the initial mention. However, if these unrelated candidates integrate well into the graph, the PPR algorithm may assign them a high score, even though they are not the correct linking decision. The impact of this error type amplifies with the number of entity mentions featuring ``imperfect" candidate lists.

Different entities require different linking approaches, hence it is essential to combine different approaches to minimize the drawbacks of each one.

\subsection{X-Linker: pipeline for named entity linking}

\begin{figure*}[!t]
    \centering
    \includegraphics[height=0.8\textwidth]{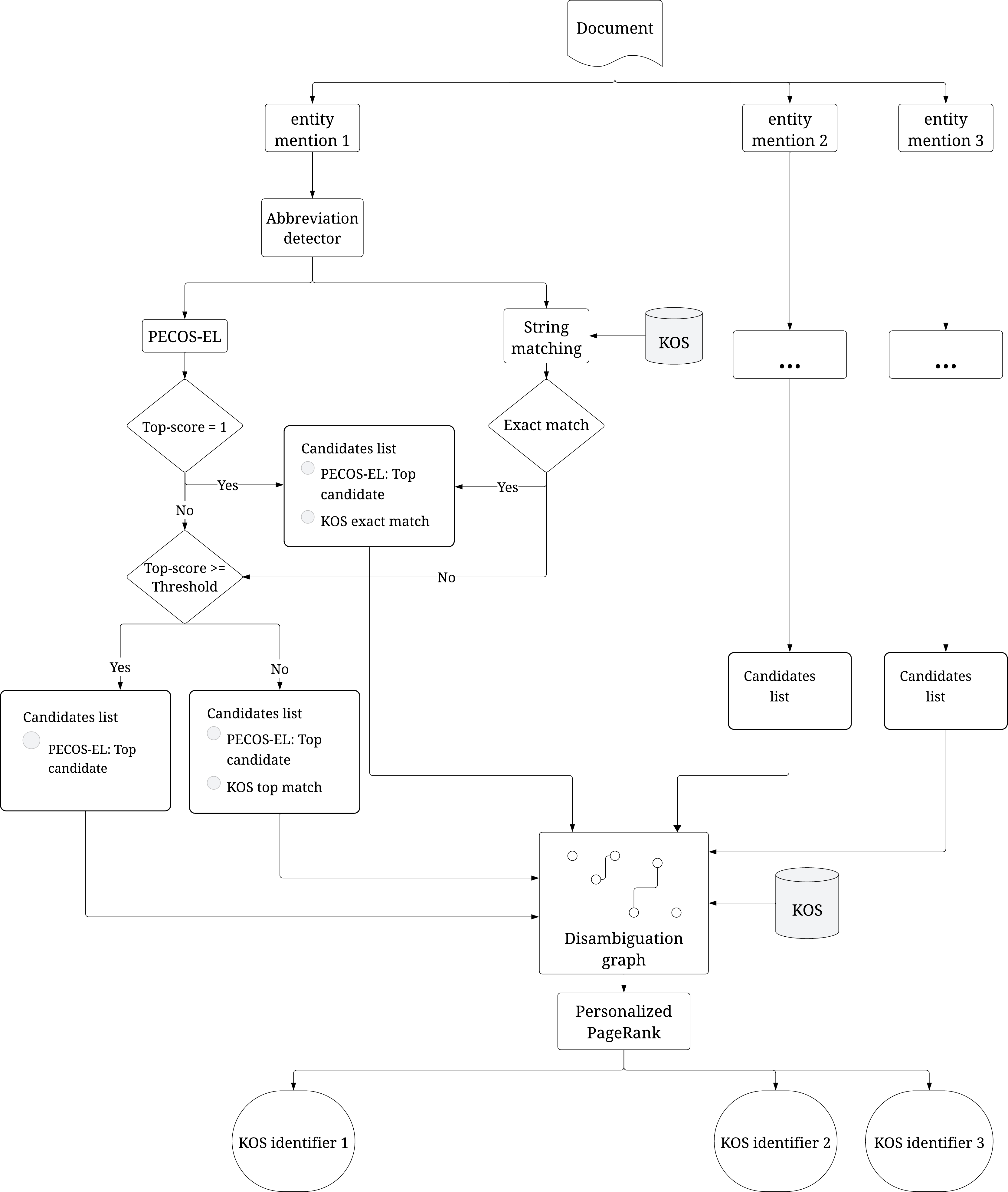}
    \caption{X-Linker pipeline to link biomedical entities to target KOS.}
   \label{fig:X_linker_pipeline}
\end{figure*}

To deal with different entities, we explore the combination of the previous approaches into a single pipeline, designated by \textit{X-Linker}. \textit{X-Linker} is a heuristic that resorts to abbreviation detection, string matching, to the PECOS-EL model and the PPR-based model according to the entity being linked. The overview of the X-Linker pipeline is shown in Fig. \ref{fig:X_linker_pipeline} and the pseudo-code is shown in Algorithm \ref{alg:x_linker}.

\begin{algorithm}
\caption{X-Linker pipeline}
\label{alg:x_linker}
\begin{algorithmic}
\STATE \textbf{Input:} $M$ 

\STATE \textbf{Initialize:} $threshold$ 

\FOR{each $m \in M$}
    \STATE $long\_m \gets$ apply\_abbreviation\_detector($m$)
    \STATE $C \gets []$ 
    \STATE $string\_matches \gets$ apply\_string\_matcher($long\_m$)
    \STATE $pecos\_matches \gets$ apply\_pecos\_el($long\_m$)
    \IF{$string\_matches$.top\_candidate['score'] $== 1.0$}
        \STATE $C$.append($string\_matches$.top\_candidate)
    \ENDIF
    \IF{$pecos\_matches$.top\_candidate['score'] $== 1.0$}
        \STATE $C$.append($pecos\_matches$.top\_candidate)
    \ELSE
        \IF{$pecos\_matches$.top\_candidate['score'] $>= threshold$}
            \STATE $C$.append($pecos\_matches$.top\_candidate)
        \ELSE
            \STATE $C$.append($pecos\_matches$.top\_candidate)
            \STATE $C$.append($string\_matches$.top\_candidate)
        \ENDIF
    \ENDIF
\ENDFOR

\STATE \textbf{Initialize:} $G$ 
\STATE $G \gets$ build\_disambiguation\_graph($C$)
\STATE $PPR\_scores \gets$ apply\_ppr\_model($G$)

\FOR{each entity mention $m \in M$}
    \STATE pick\_highest\_scoring\_candidate($PPR\_scores(m)$)
\ENDFOR
\end{algorithmic}
\end{algorithm}

The algorithm starts by taking a set of entity mentions $M$ and initializing a score threshold for filtering matches outputted by the respective PECOS-EL model. For each mention $m$, it applies an abbreviation detector to convert $m$ to its long form $long\_m$. Then, it retrieves candidate matches using a string matcher and the PECOS-EL model, storing results in $string\_matches$ and $pecos\_matches$, respectively. If the top candidate from $string\_matches$ or $pecos\_matches$ has a perfect score (1.0), it is added to the candidate list $C$. If the top candidate from $pecos\_matches$ has a score above the threshold, it is also added to $C$. If the score is below the threshold, both the top candidate from $pecos\_matches$ and the top candidate from $string\_matches$ are added to $C$. Once candidate lists for all mentions are completed, a disambiguation graph $G$ is built based on these candidates. The PPR algorithm is then applied to $G$ to compute scores for each candidate. Finally, for each mention $m$, the highest-scoring candidate from the PPR results is selected to disambiguate the mention.

A description of the implementation of the X-Linker pipeline is available in the Appendix ``Implementation".

\section{Experiments}

\begin{table*}[ht]
    \caption{Description of the evaluation datasets.}
    \label{tab:eval_datasets}
    \centering
    \begin{tabular}{c c c c c c c}
        \toprule
        \textbf{KOS} & \textbf{Entity type} & \textbf{Dataset} & \textbf{Total} & \multicolumn{2}{c}{\textbf{NIL}} & \textbf{Used} \\  
        \cmidrule{5-6}
        & & & & No ID & Obsolete & \\ 
         \midrule
        
        \multirow{3}{0pt}{MEDIC} & \multirow{3}{0pt}{Disease} & BC5CDR \cite{bc5cdr} & 4,424 & 11 & 61 & 4,352 \\
        & & BioRED \cite{biored} & 917 & 12 & 0 & 905 \\
        & & NCBI-Disease \cite{ncbidisease} & 960 & 83 & 0 & 877 \\
        
        \midrule
        
        \multirow{3}{0pt}{CTD-Chemicals} & \multirow{3}{0pt}{Chemical} & BC5CDR \cite{bc5cdr} & 5,385 & 280 & 30 & 5,075 \\
        & & BioRED \cite{biored} & 754 & 27 & 0 & 727 \\
        & & NLM-Chem \cite{nlmchem} & 11,772 & 893 & 0 & 10,879 \\
        
        \bottomrule
    \end{tabular}
\end{table*}

The datasets used for the evaluation of the several approaches are described in Table \ref{tab:eval_datasets}. We selected datasets including annotations of the type ``Disease" or ``Chemical" that are commonly used\footnote{The source for the number of citations is Google Scholar (\url{https://scholar.google.com/}) and the search was performed on June 18th, 2024}: BC5CDR (819 citations), BioRED (74 citations), NCBI-Disease (843 citations), NLM-Chem (52 citations). From each dataset, we removed the NIL annotations, including annotations associated with no KOS identifiers (whose identifier is '-1' or '-'), but also obsolete annotations, i.e., annotations with KOS identifiers that are not present in the KOS version used in our experiments. The performance of an approach in the target evaluation dataset is assessed through the calculation of the top-k accuracy, which is defined as:

\[
\text{Top-k Accuracy} = \frac{1}{N} \sum_{i=1}^{N} \mathds{1}\{y_i \in \{\hat{y}_{i,1}, \hat{y}_{i,2}, \ldots, \hat{y}_{i,k}\}\}
\]

where:
\begin{itemize}
    \item \( N \) is the total number of evaluation instances.
    \item \( y_i \) is the true KOS identifier for the \( i \)-th instance.
    \item \( \hat{y}_{i,1}, \hat{y}_{i,2}, \ldots, \hat{y}_{i,k} \) are the top \( k \) predicted identifiers (ranked by confidence) for the \( i \)-th instance.
    \item \( \mathds{1}\{ \cdot \} \) is the indicator function, which returns 1 if the true identifier \( y_i \) is among the top \( k \) predicted identifiers \( \{\hat{y}_{i,1}, \hat{y}_{i,2}, \ldots, \hat{y}_{i,k}\} \), and 0 otherwise.
\end{itemize}

Besides the baselines defined in our work, we used the state-of-the-art approach SapBERT \cite{Liu2021} for a relative comparison of the performance.

\section{Results and discussion}

\subsection{Impact of training data in the PECOS-EL model}

\begin{table}[ht]
    \centering
    \caption{Top-1 and top-5 accuracy of the PECOS-EL Disease model trained on different training datasets and applied to the evaluation datasets BC5CDR-Disease, BioRED-Disease and NCBI-Disease.}
    \label{tab:train_pecos_disease}
    \begin{tabular}{c c c c c c c}
    \toprule
         Dataset &  \multicolumn{2}{c}{BC5CDR-Disease} & \multicolumn{2}{c}{BioRED-Disease} & \multicolumn{2}{c}{NCBI-Disease}\\
         - & top-1 & top-5 & top-1 & top-5 & top-1 & top-5 \\
    \midrule 
    Disease\_KB & 0.6473	& 0.7238 & 0.6114 & 0.6681 & 0.6519	& 0.7281 \\
    Disease\_100	& 0.7682	& 0.8495  & 0.6681	& 0.8286 & 0.5961	&  0.8464 \\
    Disease\_200	& 0.7803	& 0.8787 & 0.6790	& 0.8515 &  0.6394	&  0.8294 \\
    Disease\_300	& \textbf{0.7870}	& 0.8817  & 0.6987	& 0.8515 &  0.6837	&  0.8476 \\
    Disease\_400	& 0.7746	& \textbf{0.8847} & \textbf{0.7380}	& \textbf{0.8854} &  \textbf{0.7292}	&  \textbf{0.8737} \\
    \bottomrule
    \end{tabular}
\end{table}

To assess the impact of the size of the training data and of the addition of Pubtator3 annotations, we evaluated the performance of the model PECOS-EL-Disease trained on different versions of the training data as described in Table \ref{tab:train}. Since the training dataset ``Disease-All" is large (9,497,985), we were not able to train the PECOS-EL-Disease model in this dataset due to the out-of-memory error. The performance of the Disease PECOS-EL model when applied to the different dataset versions is shown in Table \ref{tab:train_pecos_disease}. For the PECOS-EL-Disease model, incorporating Pubtator annotations into the training data enhances performance in the EL task. Moreover, as the number of Pubtator annotations increases, performance improves accordingly. However, there are some caveats. In the NCBI-Disease dataset, the addition of Pubtator annotations to the ``Disease-100" training dataset decreases the top-1-accuracy to 0.6519 from 0.5961, which was obtained training PECOS-EL-Disease in the ``Disease-KOS" dataset. Training the model in the dataset ``Disease-200" increases the top 1 accuracy to 0.6394, still below the performance of the model trained in the dataset in the ``Disease-KOS" dataset. It's only when PECOS-EL-Disease is trained in the dataset ``Disease-300" that the top-1 accuracy surpasses the baseline (0.6837). The highest top-1 accuracy is obtained when the model is trained in the dataset ``Disease-400": 0.7292. In the BioRED evaluation dataset, the performance of the PECOS-EL-Disease increases with the number of Pubtator annotations in the training data, reaching a maximum of 0.7380. In the BC5CDR-Disease dataset, the performance of the model PECOS-EL-Disease also increases with the number of Pubtator annotations in the training data, peaking when the model is trained in the dataset ``Disease-300" with a top-1-accuracy of 0.7870 and decreasing with the model training in ``Disease-400". This contradictory result may be explained by the nature of the Pubtator annotations present in the training data, more concretely, it can be attributable to the fact that there are Pubtator annotations sharing the same string, but associated with different KOS identifiers. For an explanation of this, check the next subsection \ref{overlap}. Observing the top-5 accuracy, the performance increases with the higher number of instances in the dataset. Also, the annotation performance of Pubtator3 is not 100\%, so we can safely assume that there will be errors present in the training data which further decrease the downstream performance in the evaluation of the EL task in the selected datasets.

\subsection{Is PECOS-EL a zero-shot entity linker?}\label{overlap}

\begin{table*}[htbp]
    \caption{Overview of overlapping strings with the evaluation datasets and the training data}
    \label{tab:train_overlap}
    \centering
    \begin{tabular}{c c c c c c c}
    \toprule
        \textbf{Type} & \textbf{Train instances} & \textbf{KOS concepts} & \textbf{Dataset} & \textbf{Used} & \multicolumn{2}{c}{\textbf{Overlap}} \\
        \cmidrule{6-7}
        & & & & & Train file & KOS \\
        \midrule
        \multirow{3}{*}{Disease} & 1,402,332 (Disease-200) & \multirow{3}{*}{13,292}  & BC5CDR & 4,352	& 4,116 (94.58 \%)	& 2,815 (64.68 \%) \\
        & \multirow{2}{*}{2,275,258 (Disease-400)} &  & BioRED & 905	& 868 (95.91 \%)	& 531 (58.67 \%) \\
        & &  & NCBI-Disease & 877	& 823 (93.84 \%)	& 523 (59.64 \%) \\
        \midrule
        \multirow{3}{*}{Chemical} &  \multirow{3}{*}{1,123,842} & \multirow{3}{*}{176,444}  & BC5CDR  & 5075	& 4,922 (96.99 \%)	& 4,067 (80.14 \%)  \\
        &  &  & BioRED  &  727	& 675 (92.85 \%)	& 497 (68.36 \%)\\
        & &  &  NLM-Chem &  10,879	& 9,472 (87.07 \%)	& 5,171 (47.53 \%) \\
    \bottomrule
    \end{tabular}
\end{table*}

We analysed for each evaluation dataset the percentage of annotations with strings that are also present in the data used to train the PECOS-EL model, as seen in Table \ref{tab:train_overlap}. Following the strict definition for zero-shot evaluation, i.e., an EL approach must be able to link entities that were not seen during training using only the entity descriptions, the PECOS-EL models are not zero-shot entity linkers \cite{Sevgili2022}. However, we followed the refined evaluation method recommended by the authors in \cite{Tutubalina2020}. Specifically, we removed all documents from the Pubtator set that were also present in the test sets of the evaluation datasets. The training data was gathered from the natural distribution of entities in biomedical literature. Therefore, we assume that the performance of X-Linker is robust since it is not dependent on a specific evaluation dataset. The only drawback is that the training data is biased towards the past, in the sense that is based on text already existing. There is no assurance that the same entities will continue to appear in biomedical literature in the future. However, the X-Linker approach can be updated with new training data and, in the event of new entities emerging, there is the potential to employ an approach that specifically handles NIL or unlinkable entities. This type of approach helps prevent the loss of semantic information and mitigates decreases in performance by EL approaches \cite{Ruas2022,Ran2023}.

\subsection{Impact of abbreviation detection}

\begin{table*}[ht]
    \caption{Impact of adding different modules to the X-Linker pipeline}
    \label{tab:x_linker}
    \centering
    \begin{tabular}{lcccccc}
        \toprule
        \multirow{2}{*}{Module} & \multicolumn{3}{c}{Disease} & \multicolumn{3}{c}{Chemical} \\
        \cmidrule(lr){2-4} \cmidrule(lr){5-7}
        & BC5CDR & BioRED & NCBI-Disease & BC5CDR & BioRED & NLM-Chem \\
        \midrule
        PECOS & 0.7803	& 0.7380	& 0.7292	& 0.8051	& 0.7729	& 0.6592 \\
        +abbrv\_detect & 0.8079	& 0.7664	& 0.7952	& 0.8564	& 0.8345	& 0.7164 \\
        +abbrv\_detect+SM & 0.8228	& 0.7937	& \textbf{0.8271}	& 0.9492	& \textbf{0.9248}	& 0.7850 \\
        +abbrv\_detect+SM+PPR & \textbf{0.8307}	& \textbf{0.7969}	& \textbf{0.8271}	& \textbf{0.9511}	& \textbf{0.9248}	& \textbf{0.7895} \\
        \bottomrule
    \end{tabular}
\end{table*}

As shown in Table \ref{tab:x_linker}, adding an abbreviation detection module greatly improves the performance of PECOS-EL. PECOS-EL relies solely on variations in the text of an entity for training and does not consider its context, thus its performance is highly dependent on the input mention text. \cite{Jiang2024} showed that considering the mention's context makes the approach more robust to text variations, but the resources required to train such a model leave that work for future exploration.

\subsection{Impact of string matching and the rule-based filter}

\begin{table*}[htbp]
    \caption{Overview of overlapping strings in the training and evaluation datasets and respective correctness of the associated KOS identifiers according to the evaluation datasets annotation.}
    \label{tab:train_overlap_2}
    \centering
    \begin{tabular}{ccccccc}
        \toprule
        \multirow{2}{*}{\textbf{Type}} & \multirow{2}{*}{\textbf{Dataset}} & \multicolumn{3}{c}{\textbf{Train set overlap}} & \multicolumn{2}{c}{\textbf{Correct KOS ID in the list}} \\
        
        &&  Total & Correct KOS ID in the list & Incorrect & Exact match & Ambiguous \\
        \midrule
        \multirow{3}{*}{Disease} & BC5CDR  & 4,116 (94.58 \%) &  3,779 (91.81\%) & 337 (8.92\%) & 2,743 (72.59\%) & 1,036 (27.41\%) \\
        & BioRED  & 868 (95.91 \%)  & 775 (89.29\%) & 93 (12.0\%) & 381 (49.16\%) & 394 (50.84\%)   \\
        & NCBI-Disease  &  823 (93.84 \%) & 751 (91.25\%) & 72 (9.59\%) & 362 (48.2\%) & 389 (51.8\%) \\

        \midrule

        \multirow{3}{*}{Chemical} & BC5CDR  & 4922 (96.99 \%) & 4,811 (97.74\%) &  111 (2.31\%) & 3,862 (80.27\%) & 949 (19.73\%)  \\
        & BioRED  & 675 (92.85 \%) & 665 (98.52\%) & 10 (1.5\%) & 485 (72.93\%) & 180 (27.07\%)\\
        & NLM-Chem  &  9,472 (87.07 \%) & 8,789 (92.79\%) &  683 (7.77\%) & 5,094 (57.96\%) & 3,695 (42.04\%) \\
        
        \bottomrule
    \end{tabular}
\end{table*}

Table \ref{tab:x_linker} shows the impact of adding the string matcher module to the X-Linker pipeline, which showed advantages in all evaluation datasets. Analysing the data shown in \ref{tab:train_overlap_2}, the overlap of string between training and evaluation data ranges from 96.99\% in the BC5CDR-Chemical dataset to 87.07\% in the NLM-Chem dataset. However, some of the strings in the training data are associated with more than one KOS identifier. Moreover, in some cases the identifier for a given string in the training dataset is not the same identifier associated with the same string in the evaluation dataset (check column ``Incorrect" in Table \ref{tab:train_overlap_2}). For example, 12.0\% of the strings present in the ``Disease" training dataset that are also present in the BioRED-Disease dataset are associated with different identifiers. Even if the KOS identifier that appears associated with a given string in the evaluation dataset is also associated with the same string in the training dataset, there is a relevant part of ambiguity (check column ``Ambiguous" in Table \ref{tab:train_overlap_2}), i.e., there are more than one identifier for the string. For example, in the BioRED-Disease dataset, 89.29\% of the strings in the training dataset are associated with the correct identifier as defined in the evaluation dataset, but only 49.16\% of those strings have only one identifier. The remaining 50.84\% strings have more than one associated identifier.
  
As shown in Table \ref{tab:train_overlap_2}, there are annotations in the evaluation datasets associated with entity names/strings that have an exact match in the respective target KOS. However, not always the identifier associated with the exact matching is the same as the identifier chosen to annotate the entities in the evaluation datasets. This highlights the inherent ambiguity of the annotation process, but also that the task EL does not have a universal definition. The annotation criteria are strictly associated with the scope of the motivation. For example, in the context of an annotation project centred on rare diseases, the annotation guidelines will instruct annotators to prioritise selecting more specific diseases. However, if the project encompasses various entity types simultaneously, such as chemicals, anatomical parts, cell types, etc., the annotation guidelines may not necessitate the same level of specificity as in the case of rare diseases. In such instances, a broader categorization may be sufficient to fulfil the project's objectives. Evaluation datasets are useful to straightforwardly assess the performance of EL approaches, which can be then complemented by more extensive and realistic evaluations, for example, user testing. A rule-based pipeline such as X-Linker helps diminish the impact of these disparities.

\subsection{Document context improves the performance}
Table \ref{tab:x_linker} shows the impact of adding the PPR algorithm-based module to the X-Linker pipeline. With the previously mentioned modules, the PECOS-EL module jointly with the abbreviation detector and the string matcher can deal with a large part of the entities present in the evaluation datasets. However, context in the EL task is relevant, since the same entity string can have multiple meanings according to the surrounding entities. For that, establishing a measure of coherence between a given entity and the other entities present in the same document can help to disambiguate decisions, as shown in the literature \cite{Pershina2015,Phan2019}.

\begin{figure*}[!t]
    \centering
    \includegraphics[height=0.6\textheight]{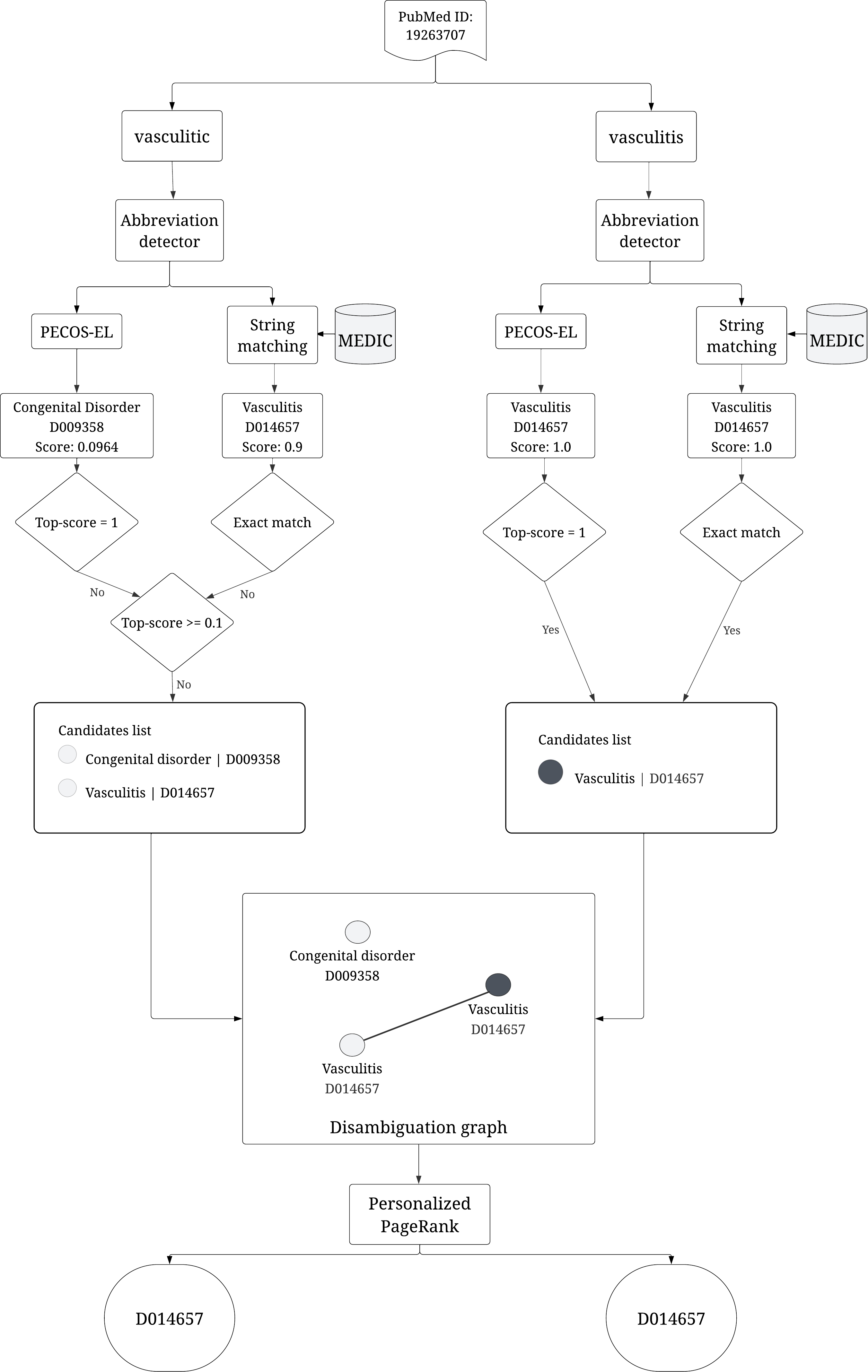}
    \caption{Example of the application of the X-Linker pipeline to the BC5CRD dataset involving the entity mentions ``vasculitis" and ``vasculitic".}
   \label{fig:example_pipeline}
\end{figure*}

Figure \ref{fig:example_pipeline} shows an example of how the X-Linker pipeline links two entity mentions present in the document with PubMed ID 19263707 from the BC5CDR-Disease dataset to entries in the MEDIC vocabulary: ``vasculitic" and ``vasculitis." As a first step, X-Linker applies abbreviation detection to each mention. Then, the model PECOS-EL-Disease predicts the candidates ``Congenital Disorder" with identifier \texttt{D009358} and ``vasculitis" with identifier \texttt{D014657} for the mentions ``vasculitic" and ``vasculitis" respectively. Concurrently, the string matcher retrieves the candidate ``vasculitis" (\texttt{D014657}) from MEDIC for both mentions. In the disambiguation process, for ``vasculitic" PECOS-EL-Disease scores low (0.0964), and the string matcher finds a close candidate (score 0.9). Both are added to the candidate list due to the low PECOS-EL-Disease score. For ``vasculitis" PECOS-EL-Disease scores 1.0, and the string matcher confirms an exact match (MEDIC). Only ``vasculitis" (\texttt{D014657}) is listed. Both lists feature ``vasculitis" linked in the disambiguation graph by MEDIC relations. The PPR selects ``vasculitis" (\texttt{D014657}) as the top candidate, resolving ambiguity.

\subsection{Comparison with SapBERT}

\begin{table*}[ht]
    \caption{Top-1 Accuracy of the X-Linker approach compared to PECOS-EL and the baseline state-of-the-art SapBERT.}
    \label{tab:sapbert}
    \centering
    \begin{tabular}{l c c c c c c}
    \toprule
         \multirow{2}{*}{\textbf{Model}} & \multicolumn{3}{c}{\textbf{Disease}} & \multicolumn{3}{c}{\textbf{Chemical}}  \\
         \cmidrule{2-7}
         & BC5CDR & BioRED & NCBI-Disease & BC5CDR & BioRED & NLM-Chem \\
         \midrule
         SapBERT & 0.7824 &  0.7434 & 0.7845 &  0.8664 & 0.7661 & 0.6678 \\
         +abbrv. detection & 0.8141 & \textbf{0.8177} & 0.8233 & \textbf{0.9559} &  0.9001 & \textbf{0.7950} \\
         \midrule
         PECOS-EL & 0.7803 & 0.7380 &	0.7292 &	0.8051 &	0.7729 &	0.6592 \\
         +abbrv. detection & 0.8079 & 0.7664	& 0.7952 & 0.8564 & 0.8345 & 0.7164 \\
        \midrule
         X-Linker (best) & \textbf{0.8307} & 0.7969 & \textbf{0.8271} &  0.9511 & \textbf{0.9248} & 0.7895 \\
         \bottomrule
    \end{tabular}
\end{table*}

The state-of-the-art EL approach SapBERT exhibits a high top-1 accuracy across all evaluated datasets, particularly for ``Chemical" entities. Like the PECOS-EL model, SapBERT relies on the mention string. Therefore, we also present its results after applying a pre-processing step of abbreviation detection for a fairer comparison. X-Linker achieves higher performance in three of the evaluation datasets: BC5CDR-Disease, NCBI-Disease and BioRED-Chemical. SapBERT's performance is higher in the remaining three evaluation datasets: BioRED-Disease, BC5CDR-Chemical and NLM-Chem. For entities of type ``Disease", SapBERT's performance is higher in a smaller dataset (BioRED-Disease), whereas for entities of type ``Chemical" SapBERT's performance is higher in the larger datasets (BC5CDR-Chemical and NLM-Chem). X-Linker's performance is higher than the performance of PECOS-EL, which highlights the importance of combining different types of EL approaches.

\subsection{Error analysis}
One type of error is related to the specificity of the annotations. For instance, the entity ``liver neoplasms" (document 26033014 in the BC5CDR dataset) is annotated with the MEDIC concept ``Liver neoplasms" (identifier \texttt{D008113}) and X-Linker correctly links the entity mention to the referred concept. However, in the same document, the entity mention `liver cancer" has the candidates ``Liver neoplasms" (\texttt{D008113}) and ``Carcinoma, hepatocellular" (\texttt{D006528}) and X-Linker links the entity mention to the child concept (\texttt{D006528}) instead of the correct one, the parent concept \texttt{D008113}. The same happens with the entity mention ``cognitive impairment" (document 24802403 in BC5CDR dataset) which X-Linker links the entity mention to the parent concept ``Cognitive dysfunction" (D060825) instead of the correct parent concept ``Cognition disorders" (D003072). This relates to the implementation of the PPR algorithm, which considers the IC of each concept to score the candidates. As a result, more specific terms are preferred over more general ones. Nevertheless, the opposite also happens: the entity mention ``Deterioration of vision" is linked to the parent concept ``vision disorders" (\texttt{D014786}) instead of the correct child concept ``Vision, Low" (\texttt{D015354}).

In other cases, the X-Linker approach is unable to produce a candidate list with the correct candidate. The entity mention ``AL"  (document 24040781 of the BC5CDR dataset) is an abbreviation of ``Amyloidosis", so it should be linked to the concept ``Amyloidosis" (\texttt{D000686}). However the generated candidates are ``Mousa Al din Al Nassar syndrome" (\texttt{C536989}), ``Pallor" (\texttt{D010167}) and Abetalipoproteinemia (\texttt{D000012}). The abbreviation detector fails to identify the abbreviation, and X-Linker generates wrong candidates. In another case, the entity mention ``mania" (document 19447152 of the BC5CDR dataset) is linked to the concept instead of the concept ``Mania" (\texttt{D000087122}) instead of the concept ``Bipolar Disorder" (\texttt{D001714}). In the Disease dataset used to train the PECOS-EL-Disease model the string ``mania" as annotated with the identifier (\texttt{D000087122}) so the model outputted this identifier.

Another type of error is related to composite mentions, since X-Linker fails to deal with these mentions. The entity mention ``hemorrhagic strokes" (document 19293073 of the BC5CDR dataset) is annotated with the identifiers \texttt{D020300} (``Intracranial Hemorrhages ") and \texttt{D020521} (``Stroke"), but X-Linker links the mention to the concept ``Hemorrhagic Stroke" (\texttt{D000083302}).

\subsection{Limitations}
There are several limitations associated with X-Linker. First, the performance of the PECOS-EL models is influenced by the accuracy of the automatic annotations provided by PubTator3 used for training. Any discrepancies arising from the automatic annotation will lead to downstream lower performance in the evaluation process using datasets. The matcher component of the PECOS-EL model uses BioBERT as the encoder model, meaning any biases associated with BioBERT may affect the results. Due to memory constraints, we did not train the PECOS-EL-Disease model on the entire training dataset and both PECOS-EL models were trained solely on the entity text, without incorporating the respective context. Training the PECOS-EL models requires significant GPU resources, so we did not perform extensive hyperparameter optimization, which may have resulted in suboptimal performance compared to fully optimized models.

\section{Conclusion}
We generated large-scale training datasets including automatic annotations to train a deep-learning-based XMR approach adapted to the biomedical EL designated by PECOS-EL. This module was integrated into the hybrid pipeline X-Linker, an EL approach including different modules to link disease and chemical entities to the MEDIC and CTD-Chemical vocabularies without the need for human-labelled data. We carried out an extensive evaluation of the X-Linker approach, resulting in top-1 accuracy values of 0.8307, 0.7969, 0.8271, 0.9511, 0.9248, 0.7895 in the datasets BC5CDR-Disease, BioRED-Disease, NCBI-Disease, BC5CDR-Chemical, BioRED-Chemical and NLM-Chem, respectively. X-Linker demonstrated superior performance compared to SapBERT in three datasets: BC5CDR-Disease, NCBI-Disease, and BioRED-Chemical. The source code is publicly available: \url{https://github.com/lasigeBioTM/X-Linker}.

In future work, we plan to enhance entity linking using X-Linker to connect mentions to the UMLS. While our current study focused on smaller KOS due to computational limits, future directions include adapting PECOS-EL to utilize the UMLS with lightweight BERT-based matchers. Additionally, we'll explore integrating NCBI Gene and Taxonomy data from Pubtator3 for generating training datasets. Currently, PECOS-EL employs a modified K-means algorithm based on string representations of KOS entities, so we aim to boost model performance by exploring different clustering approaches that incorporate KOS information and metadata.

\subsection{Acknowledgments}
\noindent This work was supported by FCT (\textit{Fundação para a Ciência e a Tecnologia}) through funding of the PhD Scholarship with ref. 2020.05393.BD, and the LASIGE Research Unit, ref. UIDB/00408/2020 (https://doi.org/10.54499/UIDB/00408/2020) and ref. UIDP/00408/2020 (https://doi.org/10.54499/UIDP/00408/2020) and by \textit{Ministerio de Ciencia e Innovación (MICINN)} under project AEI/10.13039/501100011033.

\bibliographystyle{IEEEtran}
\bibliography{references}

\newpage

\vspace{-33pt}
\begin{IEEEbiography}[{\includegraphics[width=0.9in,height=1.15in,clip,keepaspectratio]{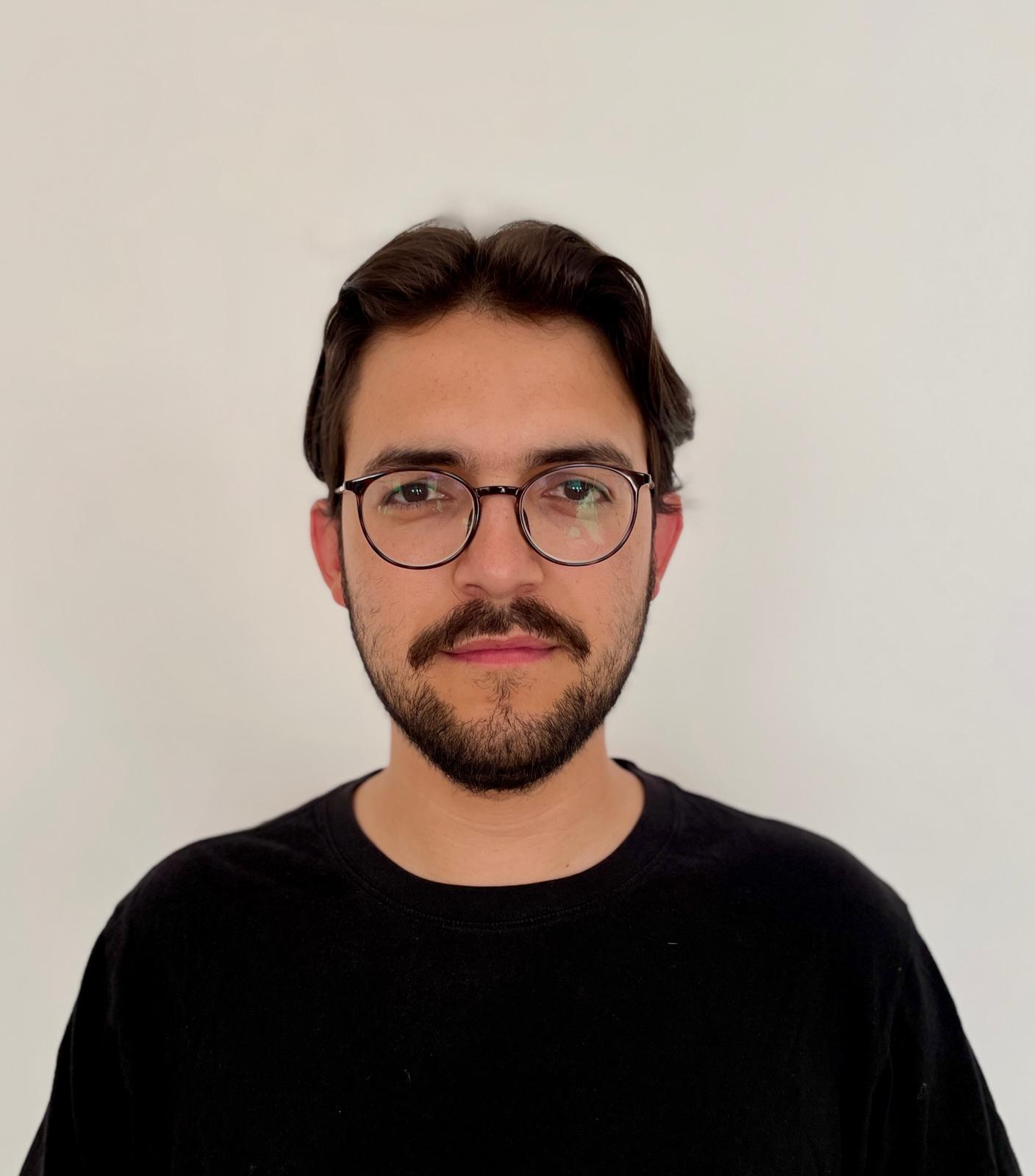}}]
{Pedro Ruas}
Pedro is a PhD candidate in Informatics at LASIGE (Faculty of Science of the University of Lisbon). He graduated in Biochemistry in 2014 from the Faculty of Science of the University of Porto and holds a Master's in Bioinformatics from the Faculty of Science of the University of Lisbon. He has been involved in projects involving natural language processing, knowledge management, and information extraction, focusing on tasks such as named entity recognition, entity linking, relation extraction, and text classification. His PhD project involves the improvement of biomedical entity linking with a focus on scientific and clinical text. He was co-author of 3 journal papers (Q1 Scimago).
\end{IEEEbiography}

\vspace{-25pt}

\begin{IEEEbiography}[{\includegraphics[width=0.9in,height=1.15in,clip,keepaspectratio]{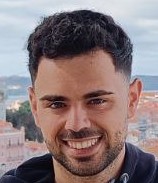}}]
{Fernando Gallego}
Fernando has received his B.Sc. degree in Software Engineering from University of Malaga, Malaga, Spain in 2019, followed by an M.Sc. degree in Artificial Intelligence with Data Science specialization from University of Malaga, Malaga, Spain in 2021. He is also a Ph.D. candidate at University of Malaga, and member of the Computational Intelligence in Biomedicine (ICB). His work focuses on extracting information present in electronic health records and leveraging it to improve decision-making in the medical sector. In this regard, natural language processing techniques such as named entity recognition or linking to different standards are essential.
\end{IEEEbiography}

\vspace{-25pt}

\begin{IEEEbiography}[{\includegraphics[width=0.9in,height=1.15in,clip,keepaspectratio]{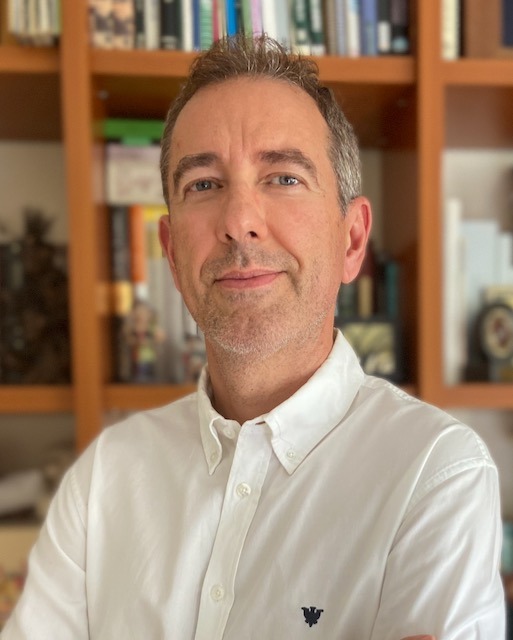}}]
{Francisco J. Veredas}
Francisco J. Veredas completed a degree in Computer Engineering at the University of Málaga in 1996 and earned a Ph.D. in Computer Engineering in 2004, focusing on computational neuroscience. He conducted predoctoral research in the USA and Germany. Since 1998, he has been an associate professor at the University of Málaga in the Department of Computer Languages and Computer Science, becoming a tenured professor in 2009. He has been involved in numerous national, European, and regional research projects, primarily applying AI techniques in biomedicine and health. From 2006 to 2015, he led the Group of Artificial Intelligence Applications in Health, achieving significant research funding and publishing impactful work on pressure ulcer diagnosis using AI. Between 2015 and 2017, he co-directed the Energy and Information in Biology group, focusing on AI for predicting protein modifications. Currently, his research centers on Natural Language Processing (NLP), co-leading projects on AI applications in clinical text processing and smart agriculture. He has supervised two doctoral thesis and is currently supervising two more, including one on clinical NLP. He is active in various research groups, is a member of the Multilingual Language Technologies Research Institute at the University of Málaga, and serves on the editorial board of Neural Computing and Applications. He has an h-index of 11 with 36 indexed publications and 440 citations.
\end{IEEEbiography}

\vspace{-25pt}
\begin{IEEEbiography}[{\includegraphics[width=0.9in,height=1.15in,clip,keepaspectratio]{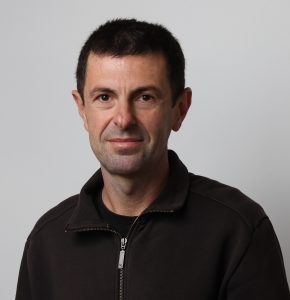}}]
{Francisco M. Couto}
Francisco M. Couto is currently an associate professor with habilitation at Universidade de Lisboa (Faculty of Sciences) and a researcher at LASIGE. He graduated (2000) and has a master (2001) in Informatics and Computer Engineering from the IST. He concluded his doctorate (2006) in Informatics, specialization Bioinformatics, from the Universidade de Lisboa. He was an invited researcher at EBI, AFMB-CNRS, BioAlma during his doctoral studies. Until 2023, he published 2 books; was co-author of 10 chapters, 65 journal papers (49 Q1 Scimago), and 33 conference papers (10 core A and A*); and was the supervisor of 11 PhD theses and of 53 master theses. At that same date, he had more than 6 thousand citations with an h-index of 37 at Google Scholar. He received the Young Engineer Innovation Prize 2004 from the Portuguese Engineers Guild, and an honorable mention in 2017 and the prize in 2018 of the ULisboa/Caixa Geral de Depósitos (CGD) Scientific Prizes. With a diverse research portfolio spanning bioinformatics, knowledge management, and information retrieval, he has contributed significantly to the advancement of various fields, including semantic similarity, ontology matching, relation extraction, and named entity recognition and linking, with a particular focus on biomedical applications. 
\end{IEEEbiography}

\vfill

\appendix[Implementation]
Our approach includes as a first step the rule-based abbreviation detector Ab3P created by \cite{Sohn2008}. To implement X-Linker we used the PECOS framework \cite{yu2022}, with the code available at \url{https://github.com/amzn/pecos}. Model training was done in two setups: (1) a server including 2 Intel(R) Xeon(R) Silver 4114 CPU @ 2.20GHz and 8 Tesla M10 GPUs; Total CPU memory: $\approx$ 64 GB. Total GPU memory: $\approx$ 64 GB; (2) an HPC cluster with 8 nodes with each 2 x AMD EPYC 7742 processors/node x 64 cores 128 cores, 1024 GB RAM, 40 GB VRAM each GPU, 4 GPU NVIDIA A100 (only 80GB RAM were used for training). Training time varied according to the entity type and the number of instances: Disease-400 (the large file with Disease entities) $\approx$ 8 hours in the HPC cluster; Chemical $\approx$ 16 hours. In the X-Linker pipeline, the threshold for candidate filtering is set to 0.1 as the default. 

\end{document}